\documentclass[11pt,a4paper]{article}
\usepackage[hyperref]{emnlp-ijcnlp-2019}
\usepackage{times}
\usepackage{latexsym}
\usepackage{xcolor}
\usepackage{multirow}
\usepackage{graphicx}
\usepackage{subfig}
\usepackage{booktabs}

\usepackage{url}

\aclfinalcopy %

\title{Findings of the Third Workshop on \\ Neural Generation and Translation}

\author{Hiroaki Hayashi$^{\diamondsuit}$, Yusuke Oda$^{\clubsuit}$, Alexandra Birch$^{\spadesuit}$, Ioannis Konstas$^{\bigtriangleup}$, \\
        \textbf{Andrew Finch$^{\heartsuit}$, Minh-Thang Luong$^{\clubsuit}$, Graham Neubig$^{\diamondsuit}$, Katsuhito Sudoh$^{\star}$} \\ \\
  $^\diamondsuit$Carnegie Mellon University,
  $^\clubsuit$Google Brain,
  $^\spadesuit$University of Edinburgh \\
  $^\bigtriangleup$Heriot-Watt University,
  $^\heartsuit$Apple,
  $^\star$Nara Institute of Science and Technology
}

\date{}

\begin{document}
\maketitle
\begin{abstract}
  This document describes the findings of the Third Workshop on Neural Generation and Translation, held in concert with the annual conference of the Empirical Methods in Natural Language Processing (EMNLP 2019).
  First, we summarize the research trends of papers presented in the proceedings.
  Second, we describe the results of the two shared tasks 1) efficient neural machine translation (NMT) where participants were tasked with creating NMT systems that are both accurate and efficient, and  2) document-level generation and translation (DGT) where participants were tasked with developing systems that generate summaries from structured data, potentially with assistance from text in another language.
\end{abstract}

\section{Introduction}
\label{sec:intro}

Neural sequence to sequence models \cite{kalchbrenner13rnntm,sutskever14sequencetosequence,bahdanau15alignandtranslate} are now a workhorse behind a wide variety of different natural language processing tasks such as machine translation, generation, summarization and simplification.
The 3rd Workshop on Neural Machine Translation and Generation (WNGT 2019) provided a forum for research in applications of neural models to machine translation and other language generation tasks (including summarization \cite{rush2015neuralattention}, NLG from structured data \cite{wen2015conditioned}, dialog response generation \cite{vinyals2015neural}, among others).
Overall, the workshop was held with two goals.

First, it aimed to synthesize the current state of knowledge in neural machine translation and generation: this year we continued to encourage submissions that not only advance the state of the art through algorithmic advances, but also analyze and understand the current state of the art, pointing to future research directions.
Towards this goal, we received a number of high-quality research contributions on both workshop topics, as summarized in Section \ref{sec:researchprogram}.

Second, the workshop aimed to expand the research horizons in NMT: we continued to organize the Efficient NMT task
which encouraged participants to develop not only accurate but computationally efficient systems.
In addition, we organized a new shared task on ``Document-level Generation and Translation'', which aims to push forward document-level generation technology and contrast the methods for different types of inputs.
The results of the shared task are summarized in Sections \ref{sec:sharedtask_dgt} and \ref{sec:sharedtask_efficiency}.

\section{Summary of Research Contributions}
\label{sec:researchprogram}

We published a call for long papers, extended abstracts for preliminary work, and cross-submissions of papers submitted to other venues. The goal was to encourage discussion and interaction with researchers from related areas.

We received a total of 68 submissions, from which we accepted 36. There were three cross-submissions, seven long abstracts and 26 full papers. There were also seven system submission papers. All research papers were reviewed twice through a double blind review process, and avoiding conflicts of interest.

There were 22 papers with an application to generation of some kind, and 14 for translation which is a switch from previous workshops where the focus was on machine translation.
The caliber of the publications was very high and the number has more than doubled from last year (16 accepted papers from 25 submissions).

\section{Shared Task: Document-level Generation and Translation}
\label{sec:sharedtask_dgt}

The first shared task at the workshop focused on document-level generation and translation.
Many recent attempts at NLG have focused on sentence-level generation \cite{lebret2016neural,gardent2017webnlg}.
However, real world language generation applications tend to involve generation of much larger amount of text such as dialogues or multi-sentence summaries.
The inputs to NLG systems also vary from structured data such as tables \cite{lebret2016neural} or graphs \cite{wang2018describing}, to textual data \cite{nallapati2016abstractive}.
Because of such difference in data and domain, comparison between different methods has been nontrivial.
This task aims to (1) push forward such document-level generation technology by providing a testbed, and (2) examine the differences between generation based on different types of inputs including both structured data and translations in another language.

\begin{figure}[t]
    \centering
    \includegraphics[width=0.9\linewidth]{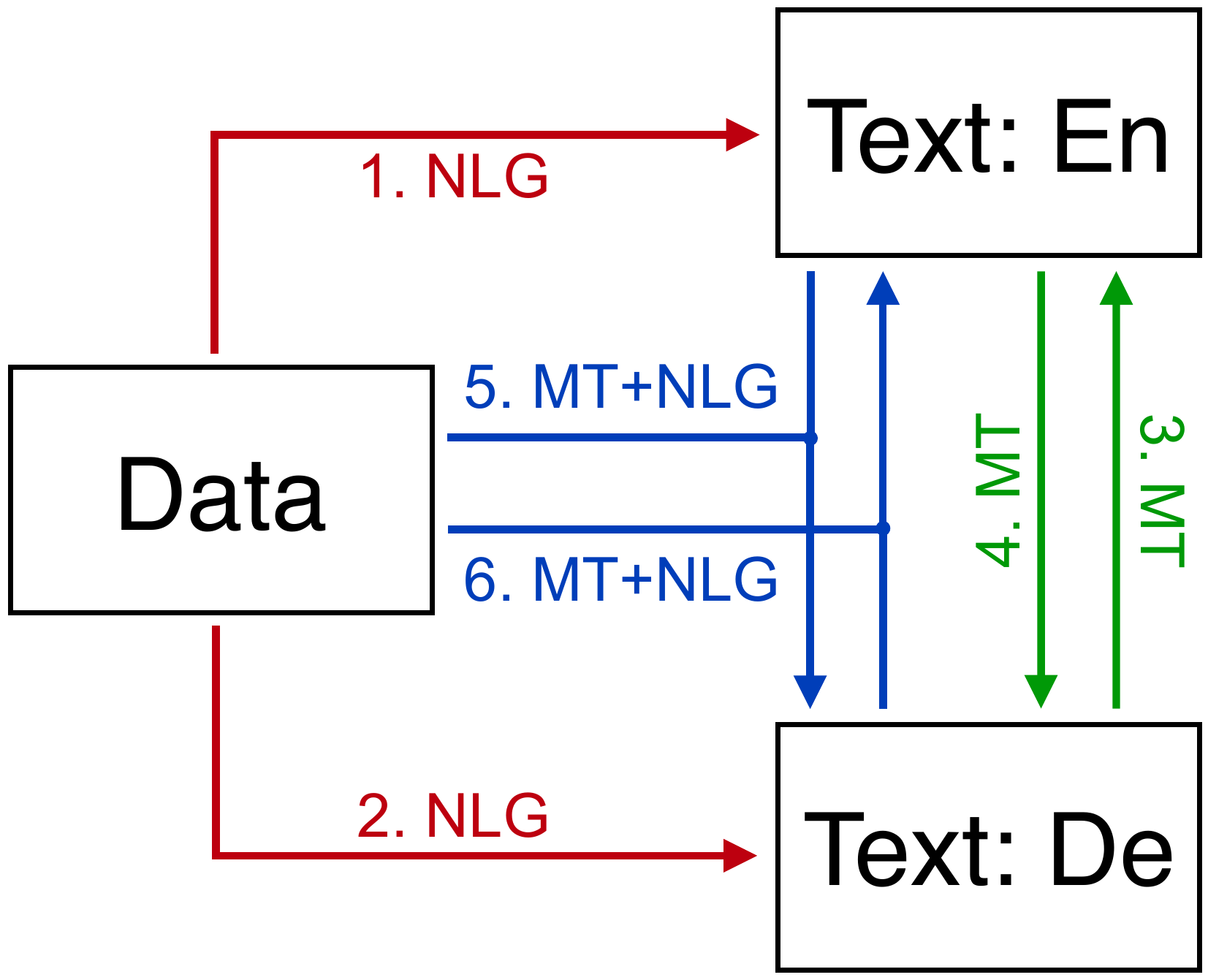}
    \caption{Input/output relationship for each track in the DGT task.}%
    \label{fig:dgt_diag}
\end{figure}

In particular, we provided the following six tracks which focus on different input/output requirements:
\begin{itemize}
    \item \textbf{NLG (Data $\rightarrow$ En, Data $\rightarrow$ De):} Generate document summaries in a target language given only structured data.
    \item \textbf{MT (De $\leftrightarrow$ En):} Translate documents in the source language to the target language.
    \item \textbf{MT+NLG (Data+En $\rightarrow$ De, Data+De $\rightarrow$ En):} Generate document summaries given the structured data and the summaries in another language.
\end{itemize}
Fig~\ref{fig:dgt_diag} shows the task diagram, where each edge represents one of the six tracks.

\subsection{Evaluation Measures}
\label{sec:eval_dgt}
We employ standard evaluation metrics for data-to-text NLG and MT along two axes:

\begin{description}
    \item[Textual Accuracy Measures:] We used BLEU \cite{papineni02bleu} and ROUGE \cite{lin2004rouge}  as measures for texutal accuracy compared to reference summaries.
    \item[Content Accuracy Measures:] We evaluate the fidelity of the generated content to the input data using relation generation (RG), content selection (CS), and content ordering (CO) metrics \cite{wiseman2017challenges}.
\end{description}

The content accuracy measures were calculated using information extraction models trained on respective target languages.
We followed \cite{wiseman2017challenges} and ensembled six information extraction models (three CNN-based, three LSTM-based) with different random seeds for each language.

\subsection{Data}
\label{sec:data_dgt|}
Due to the lack of a document-level parallel corpus which provides structured data for each instance, we took an approach of translating an existing NLG dataset.
Specifically, we used a subset of the RotoWire dataset \cite{wiseman2017challenges} and obtained professional German translations, which are sentence-aligned to the original English articles.
The obtained parallel dataset is called the RotoWire English-German dataset, and consists of box score tables, an English article, and its German translation for each instance.
Table \ref{tab:data_rotowire} shows the statistics of the obtained dataset.
We used the test split from this dataset to calculate the evaluation measures for all the tracks.

\begin{table}[t]
\centering
\begin{tabular}{lrrr}
\toprule
                    & Train & Valid & Test\\ \midrule
\# documents        & 242   & 240   & 241\\
Avg. \# tokens (En) & 323   & 328   & 329\\
Avg. \# tokens (De) & 320   & 324   & 325\\
Vocabulary size (En)& 4163  & -     & -  \\
Vocabulary size (De)& 5425  & -     & -  \\
\bottomrule

\end{tabular}
\caption{Data statistics of RotoWire English-German Dataset.}
\label{tab:data_rotowire}
\end{table}

We further allowed the following additional resources for each track:

\begin{itemize}
    \item NLG: RotoWire, Monolingual
    \item MT: WMT19, Monolingual
    \item MT+NLG: RotoWire, WMT19, Monolingual
\end{itemize}

RotoWire refers to the RotoWire dataset \cite{wiseman2017challenges} (train/valid), WMT19 refers to the
set of parallel corpora allowable by the WMT 2019 English-German task, and Monolingual refers to monolingual data allowable by the same WMT 2019 task, pre-trained embeddings (\textit{e.g.}, GloVe \cite{pennington2014glove}), pre-trained contextualized embeddings (\textit{e.g.}, BERT \cite{devlin2019bert}), pre-trained language models (\textit{e.g.}, GPT-2 \cite{radford2019language}).

Systems which follow these resource constraints are marked constrained, otherwise unconstrained. Results are indicated by the initials (C/U).

\subsection{Baseline Systems}
Considering the difference in inputs for MT and NLG tracks, we prepared two baselines for respective tracks.

\begin{description}
    \item[FairSeq-19] FairSeq \cite{ng2019facebook} was used for MT and MT+NLG tracks for both directions of
        translations. We used the published WMT'19 single model and did not tune on in-domain data.
    \item[NCP+CC:] A two-stage model from \cite{puduppully2019data} was used for NLG tracks. We utilized the
        pretrained English model trained on RotoWire dataset for English article generation, while
        the German model was trained on RotoWire English-German dataset.
\end{description}

\subsection{Submitted Systems}
Four teams, Team EdiNLG, Team FIT-Monash, Team Microsoft, Team Naver Labs Europe, and Team SYSTRAN-AI participated in the shared task.
We note the common trends across many teams and discuss the systems of individual teams below.
On MT tracks, all the teams have adopted a variant of Transformer~\cite{vaswani2017attention} as a sequence transduction model and trained on corpora with different data-augmentation methods.
Trained systems were then fine-tuned on in-domain data including our RotoWire English-German dataset.
The focus of data augmentation was two-fold: 1) acquiring in-domain data and 2) utilizing document boundaries from existing corpora.
Most teams applied back-translation on various sources including NewsCrawl and the original RotoWire dataset for this purpose.

NLG tracks exhibited a similar trend for the sequence model selection, except for Team EdiNLG who employed LSTM.

\subsubsection{Team EdiNLG}
Team EdiNLG built their NLG system upon \cite{puduppully2019data} by extending it to
further allow copying from the table in addition to generating from vocabulary and the content plan.
Additionally, they included features indicating the win/loss team records and team rank
in terms of points for each player.
They trained the NLG model for both languages together, using a shared BPE vocabulary obtained from target game summaries and by prefixing the target text with the target language indicator.

For MT and MT+NLG tracks, they mined the in-domain data by extracting basketball-related texts from \textit{Newscrawl} when one of the following conditions are met: 1) player names from the RotoWire English-German training set appear, 2) two NBA team names appear in the same document, or 3) ``\textit{NBA}'' appears in titles. This resulted in 4.3 and 1.1 million monolingual sentences for English and German, respectively. The obtained sentences were then back-translated and added to the training corpora.
They submitted their system \textbf{EdiNLG} in all six tracks.

\subsubsection{Team FIT-Monash}
Team FIT-Monash built a document-level NMT system \cite{maruf2019selective} and participated in MT tracks.
The document-level model was initialized with a pre-trained sentence-level NMT model on news domain parallel corpora.
Two strategies for composing document-level context were proposed: flat and hierarchical attention.
Flat attention was applied on all the sentences, while hierarchical attention was computed at sentence and word-level in a hierarchical manner. Sparse attention was applied at sentence-level in order to identify key sentences that are important for translating the current sentence.

To train a document-level model, the team focused on corpora that have document boundaries,
including News Commentary, Rapid, and the RotoWire dataset. Notably, greedy decoding was employed due to computational cost.
The submitted system is an ensemble of three runs indicated as \textbf{FIT-Monash}.

\subsubsection{Team Microsoft}
Team Microsoft (MS) developed a Transformer-based NLG system which consists of two sequence-to-sequence models. The two step method was inspired by the approach from \cite{puduppully2019data}, where the first model is a recurrent pointer network that selects encoded records, and the second model takes the selected
content representation as input and generates summaries.
The proposed model (\textbf{MS-End-to-End}) learned both models at the same time with a combined loss function.
Additionally, they have investigated the use of pre-trained language models for NLG track. Specifically, they fine-tuned GPT-2 \cite{radford2019language} on concatenated pairs of (template, target) summaries, while constructing templates following \cite{wiseman2017challenges}. The two sequences are concatenated around a special token which indicates ``\textit{rewrite}''. At decoding time, they adopted nucleus sampling \cite{holtzman2019curious} to enhance the generation quality. Different thresholds for
nucleus sampling were investigated, and two systems with different thresholds were submitted:
\textbf{MS-GPT-50} and \textbf{MS-GPT-90}, where the numbers refer to Top-$p$ thresholds.

The generated summaries in English using the following systems were then translated with the MT systems
which is described below. Hence, this marks Team Microsoft's German NLG (Data $\rightarrow$ De) submission unconstrained, due to the usage of parallel data beyond the RotoWire English-German dataset.

As for the MT model, a pre-trained system from \cite{xia2019microsoft} was fine-tuned on the RotoWire English-German dataset, as well as back-translated sentences from the original RotoWire dataset for the English-to-German track.
Back-translation of sentences obtained from \textit{Newscrawl} according to the similarity to RotoWire data \cite{moore2010intelligent} was attempted but did not lead to improvement.
The resulting system is shown as \textbf{MS} on MT track reports.

\subsubsection{Team Naver Labs Europe}
Team Naver Labs Europe (NLE) took the approach of transferring the model from MT to NLG.
They first trained a sentence-level MT model by iteratively extend the training set
from the WMT19 parallel data and RotoWire English-German dataset to back-translated \textit{Newscrawl} data. The best sentence-level model was then fine-tuned at document-level, followed by fine-tuning
on the RotoWire English-German dataset (constrained \textbf{NLE}) and additionally on the back-translated original RotoWire dataset (unconstrained \textbf{NLE}).

To fully leverage the MT model, input record values prefixed with special tokens for record types were sequentially fed in a specific order.
Combined with the target summary, the pair of record representations and the target summaries formed data for a sequence-to-sequence model.
They fine-tuned their document-level MT model on these NLG data which included the original RotoWire and RotoWire English-German dataset.

The team tackled MT+NLG tracks by concatenating source language documents and the sequence of records as inputs.
To encourage the model to use record information more, they randomly masked certain portion of tokens in the source language documents.

\subsubsection{Team SYSTRAN-AI}
Team SYSTRAN-AI developed their NLG system based on the Transformer~\cite{vaswani2017attention}.
The model takes as input each record from the box score featurized into embeddings and decode
the summary. In addition, they introduced a content selection objective where the model learns
to predict whether or not each record is used in the summary, comprising a sequence of binary
classfication decision.

Furthermore, they performed data augmentation by synthesizing records whose numeric values were
randomly changed in a way that does not change the win / loss relation and remains within
a sane range. The synthesized records were used to generate a summary to obtain new (record, summary) pairs and were included added the training data.
To bias the model toward generating more records, they further fine-tuned their model on
a subset of training examples which contain $N (=16)$ records in the summary.
The submitted systems are \textbf{SYSTRAN-AI} and \textbf{SYSTRAN-AI-Detok}, which differ in
tokenization.

\subsection{Results}

We show the results for each track in Table~\ref{tab:nlg-1} through \ref{tab:mtnlg-6}.
In the NLG and MT+NLG tasks, we report BLEU, ROUGE (F1) for textual accuracy, RG (P), CS(P, R), and CO (DLD) for content accuracy.
While for MT tasks, we only report BLEU. We summarize the shared task results for each track below.

In NLG (En) track, all the participants encouragingly submitted systems outperforming a strong
baseline by \cite{puduppully2019data}.
We observed an apparent difference between the constrained and unconstrained settings.
Team NLE's approach showed that pre-training of the document-level generation model on news corpora
is effective even if the source input differs (German text vs linearized records).
Among constrained systems, it is worth noting that all the systems but Team EdiNLG used the Transformer,
but the result did not show noticeable improvements compared to EdiNLG.
It was also shown that the generation using pre-trained language models is sensitive to how the sampling
is performed; the results of MS-GPT-90 and MS-GPT-50 differ only in the nucleus sampling hyperparameter,
which led to significant differences in every evaluation measure.

The NLG (De) track imposed a greater challenge compared to its English counterpart due to the lack of training data.
The scores has generally dropped compared to NLG (En) results.
To alleviate the lack of German data, most teams developed systems under unconstrained setting by utilizing
MT resources and models.
Notably, Team NLE's has achieved similar performance to the constrained system results on NLG (En).
However, Team EdiNLG achieved similar performance under the constrained setting by fully leveraging the original
RotoWire using the sharing of vocabulary.

In MT tracks, we see the same trend that the system under unconstrained setting (NLE) outperformed all
the systems under the constrained setting. The improvement observed in the unconstrained setting came from
fine-tuning on the back-translated original RotoWire dataset, which offers purely in-domain parallel documents.

While the results are not directly comparable due to different hyperparameters used in systems,
fine-tuning on in-domain parallel sentences was shown effective (FairSeq-19 vs others).
When incorporating document-level data, it was shown that document-level models (NLE, FIT-Monash, MS) perform better than sentence-level models (EdiNLG, FairSeq-19),
even if a sentence-level model is trained on document-aware corpora.

For MT+NLG tracks, interestingly, no teams found the input structured data useful, thus applying MT models
for MT+NLG tracks.
Compared to the baseline (FairSeq-19), fine-tuning on in-domain data resulted in better performance overall as seen in the results of Team MS and NLE.
The key difference between Team MS and NLE is the existence of document-level fine-tuning, where
Team NLE outperformed in terms of textual accuracy (BLEU and ROUGE) overall, in both target languages.

\begin{table*}
\centering
\begin{tabular}{lccccccccc}
\toprule
System & BLEU & R-1 & R-2 & R-L & RG & \multicolumn{2}{c}{CS (P/R)} & CO & Type \\
\midrule
EdiNLG     & 17.01 &  44.87 &  18.53 & 25.38 & 91.41 & 30.91 & 64.13 & 21.72 & C\\
MS-GPT-90   & 13.03 &  45.25 &  15.17 & 21.34 & 88.70 & 32.84 & 50.58 & 17.36 & C\\
MS-GPT-50   & 15.17 &  45.34 &  16.64 & 22.93 & 94.35 & 33.91 & 53.82 & 19.30 & C \\
MS-End-to-End & 15.03 &  43.84 &  17.49 & 23.86 & 93.38 & 32.40 & 58.02 & 18.54 & C \\
NLE        & 20.52 &  49.38 &  22.36 & 27.29 & 94.08 & 41.13 & 54.20 & 25.64 & U \\
SYSTRAN-AI & 17.59 & 47.76 & 20.18 &  25.60 & 83.22 & 31.74 & 44.90 & 20.73 & C\\
SYSTRAN-AI-Detok & 18.32 & 47.80 & 20.19 & 25.61 & 84.16 & 34.88 & 43.29 & 22.72 & C\\\midrule
NCP+CC     & 15.80 &  44.83 &  17.07 & 23.46 & 88.59 & 30.47 & 55.38 & 18.31 & C\\
\bottomrule
\end{tabular}
\caption{Results on the NLG: (Data $\rightarrow$ \textbf{En}) track of DGT task.}
\label{tab:nlg-1}
\end{table*}

\begin{table*}
\centering
\begin{tabular}{lccccccccc}
\toprule
System & BLEU & R-1 & R-2 & R-L & RG & \multicolumn{2}{c}{CS (P/R)} & CO & Type \\
\midrule
EdiNLG        & 10.95 & 34.10 & 12.81 & 19.70 & 70.23 & 23.40 & 41.83 & 16.06 & C \\
MS-GPT-90     & 10.43 & 41.35 & 12.59 & 18.43 & 75.05 & 31.23 & 41.32 & 16.32 & U \\
MS-GPT-50     & 11.84 & 41.51 & 13.65 & 19.68 & 82.79 & 34.81 & 42.51 & 17.12 & U \\
MS-End-to-End & 11.66 & 40.02 & 14.36 & 20.67 & 80.30 & 28.33 & 49.13 & 16.54 & U \\
NLE           & 16.13 & 44.27 & 17.50 & 23.09 & 79.47 & 29.40 & 54.31 & 20.62 & U \\\midrule
NCP+CC        & \phantom{0}7.29  & 29.56 & \phantom{0}7.98  & 16.06 & 49.69 & 21.61 & 26.14 & 11.84 & C \\
\bottomrule
\end{tabular}
\caption{Results on the NLG: (Data $\rightarrow$ \textbf{De}) track of DGT task.}
\label{tab:nlg-2}
\end{table*}

\begin{table}
\centering
\begin{tabular}{lcc}
\toprule
System & BLEU & Type \\
\midrule
FIT-Monash  & 47.39 & C\\
EdiNLG      & 41.15 & C\\
MS          & 57.99 & C\\
NLE         & 62.16 & U\\
NLE         & 58.22 & C\\\midrule
FairSeq-19  & 42.91 & C\\
\bottomrule
\end{tabular}
\caption{DGT results on the MT track (De $\rightarrow$ En).}
\label{tab:mt-3}
\end{table}

\begin{table}
\centering
\begin{tabular}{lcc}
\toprule
System & BLEU & Type \\
\midrule
FIT-Monash & 41.46 & C\\
EdiNLG     & 36.85 & C\\
MS         & 47.90 & C\\
NLE        & 48.02 & C\\\midrule
FairSeq-19 & 36.26 & C\\
\bottomrule
\end{tabular}
\caption{DGT results on the MT track (En $\rightarrow$ De)}
\label{tab:mt-4}
\end{table}

\begin{table*}
\centering
\begin{tabular}{lccccccccc}
\toprule
System & BLEU & R-1 & R-2 & R-L & RG & \multicolumn{2}{c}{CS (P/R)} & CO & Type \\
\midrule
EdiNLG     & 36.85 & 69.66 & 41.47 & 57.25 & 81.01 & 77.32 & 78.49 & 62.21 & C \\
MS         & 47.90 & 75.95 & 51.75 & 65.61 & 80.98 & 76.88 & 84.57 & 67.84 & C \\
NLE        & 48.24 & 75.89 & 51.80 & 65.90 & 80.65 & 75.10 & 88.72 & 69.17 & C \\ \midrule
FairSeq-19 & 36.26 & 68.22 & 40.31 & 56.38 & 81.64 & 77.67 & 75.82 & 60.83 & C \\
\bottomrule
\end{tabular}
\caption{Results on the MT+NLG: (Data+En $\rightarrow$ \textbf{De}) track of DGT task.}
\label{tab:mtnlg-5}
\end{table*}

\begin{table*}
\centering
\begin{tabular}{lccccccrcc}
\toprule
System & BLEU & R-1 & R-2 & R-L & RG & \multicolumn{2}{c}{CS (P/R)} & CO & Type \\
\midrule
EdiNLG     & 41.15 & 76.57 & 50.97 & 66.62 & 91.40 & 78.99 & 63.04 & 51.73 &C \\
MS         & 57.99 & 83.03 & 63.03 & 75.44 & 95.77 & 92.49 & 91.62 & 84.70 &C \\
NLE        & 62.24 & 84.38 & 66.11 & 77.17 & 95.63 & 91.71 & 92.69 & 85.05 &C \\\midrule
FairSeq-19 & 42.91 & 77.57 & 52.66 & 68.66 & 93.53 & 83.33 & 84.22 & 70.47 &C \\
\bottomrule
\end{tabular}
\caption{Results on the MT+NLG: (Data+De $\rightarrow$ \textbf{En}) track of DGT task.}
\label{tab:mtnlg-6}
\end{table*}

\section{Shared Task: Efficient NMT}
\label{sec:sharedtask_efficiency}

The second shared task at the workshop focused on efficient neural machine translation.
Many MT shared tasks, such as the ones run by the Conference on Machine Translation \cite{bojar2017wmt}, aim to improve the state of the art for MT with respect to accuracy: finding the most accurate MT system regardless of computational cost.
However, in production settings, the efficiency of the implementation is also extremely important.
The efficiency shared task for WNGT (inspired by the ``small NMT'' task at the Workshop on Asian Translation \cite{nakazawa2017wat}) was focused on creating systems for NMT that are not only accurate, but also efficient.
Efficiency can include a number of concepts, including memory efficiency and computational efficiency.
This task concerns itself with both, and we cover the detail of the evaluation below.

\subsection{Evaluation Measures}
\label{sec:eval_efficiency}

We used metrics to measure several different aspects connected to how good the system is.
These were measured for systems that were run on CPU, and also systems that were run on GPU.

\begin{description}
\item[Accuracy Measures:] As a measure of translation accuracy, we used BLEU \cite{papineni02bleu} and NIST \cite{doddington02nist} scores.
\item[Computational Efficiency Measures:] We measured the amount of time it takes to translate the entirety of the test set on CPU or GPU. Time for loading models was measured by having the model translate an empty file, then subtracting this from the total time to translate the test set file.
\item[Memory Efficiency Measures:] We measured: (1) the size on disk of the model, (2) the number of parameters in the model, and (3) the peak consumption of the host memory and GPU memory.
\end{description}

These metrics were measured by having participants submit a container for the virtualization environment Docker\footnote{\url{https://www.docker.com/}}, then measuring from outside the container the usage of computation time and memory.
All evaluations were performed on dedicated instances on Amazon Web Services\footnote{\url{https://aws.amazon.com/}}, specifically of type \texttt{m5.large} for CPU evaluation, and \texttt{p3.2xlarge} (with a NVIDIA Tesla V100 GPU).

\subsection{Data}

The data used was from the WMT 2014 English-German task \cite{bojar14wmt}, using the preprocessed corpus provided by the Stanford NLP Group\footnote{\url{https://nlp.stanford.edu/projects/nmt/}}.
Use of other data was prohibited.

\begin{figure*}[t!]
  \centering
  \subfloat[CPU Time vs. Accuracy]{\includegraphics[width=0.49\textwidth]{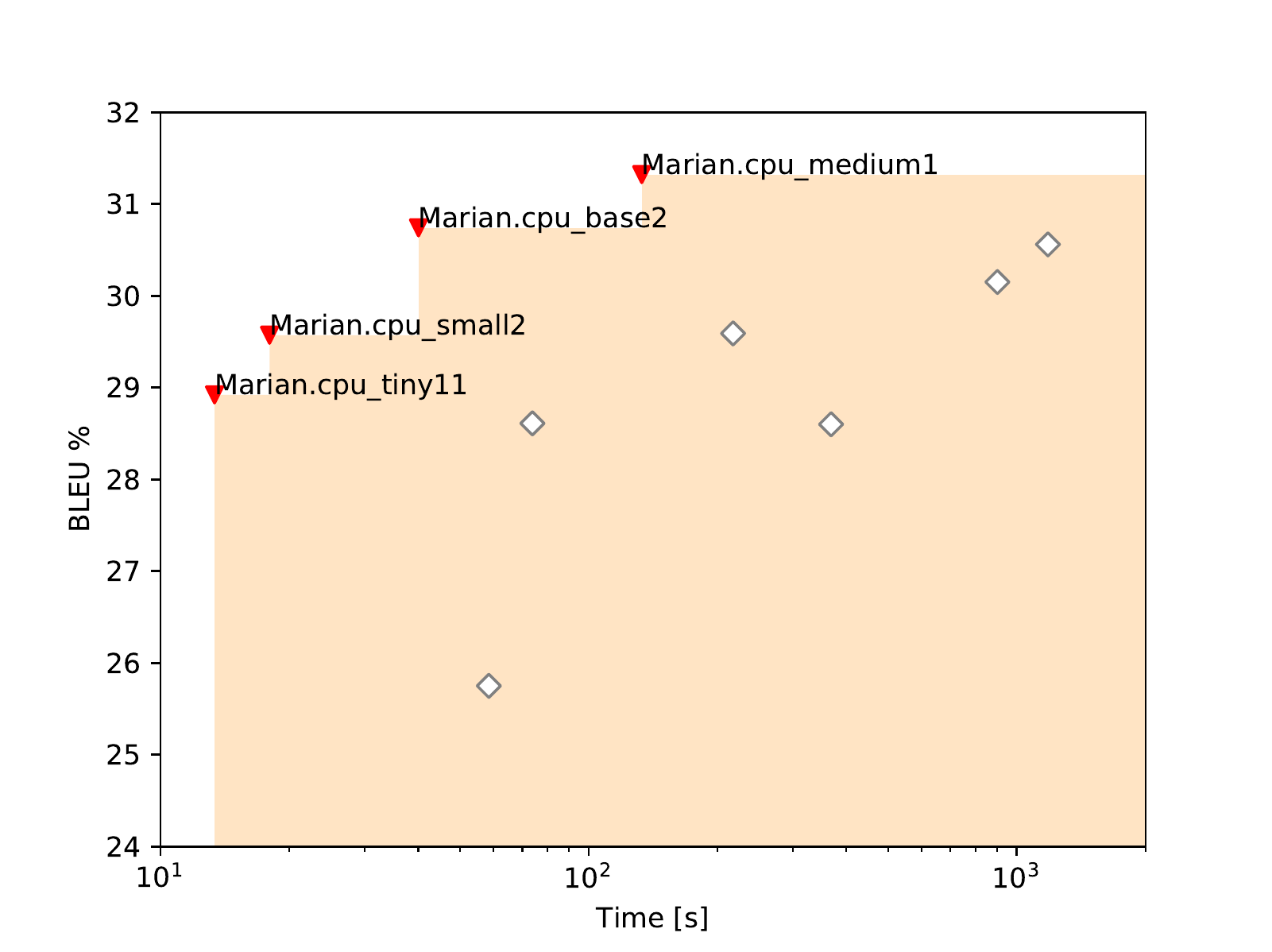} }
  \subfloat[GPU Time vs. Accuracy]{\includegraphics[width=0.49\textwidth]{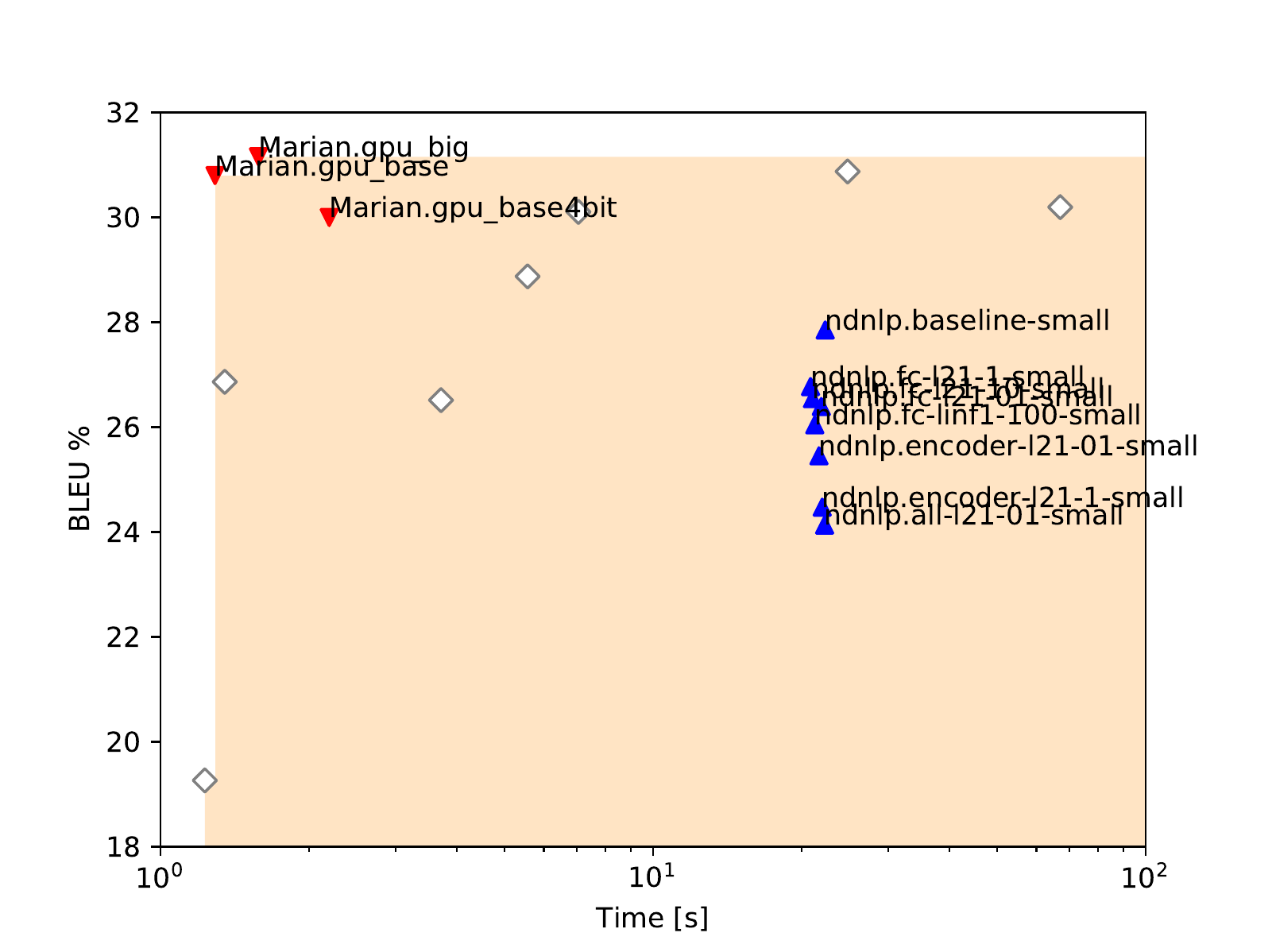} } \\
  \subfloat[CPU Memory vs. Accuracy]{\includegraphics[width=0.49\textwidth]{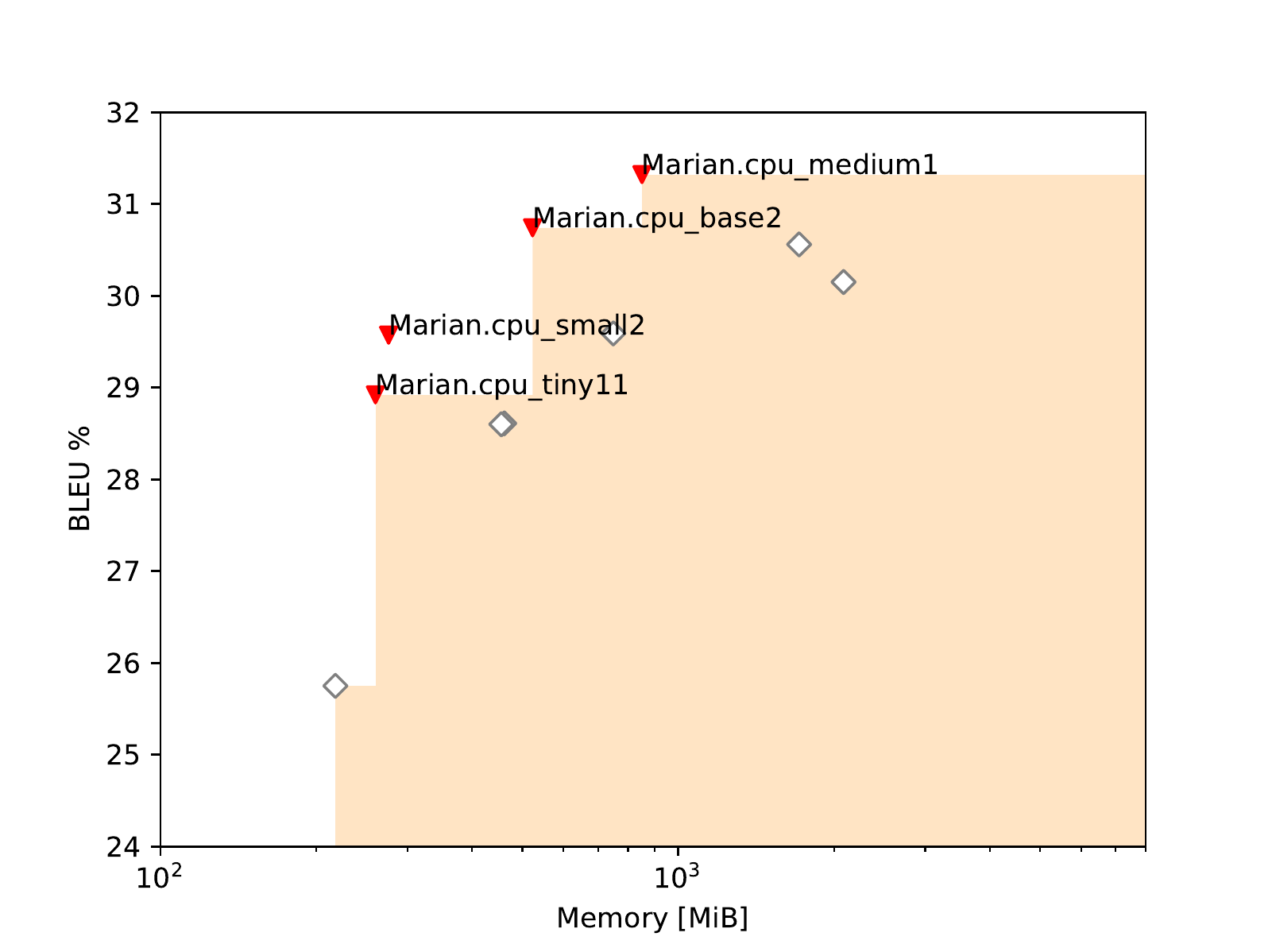} }
  \subfloat[GPU Memory vs. Accuracy]{\includegraphics[width=0.49\textwidth]{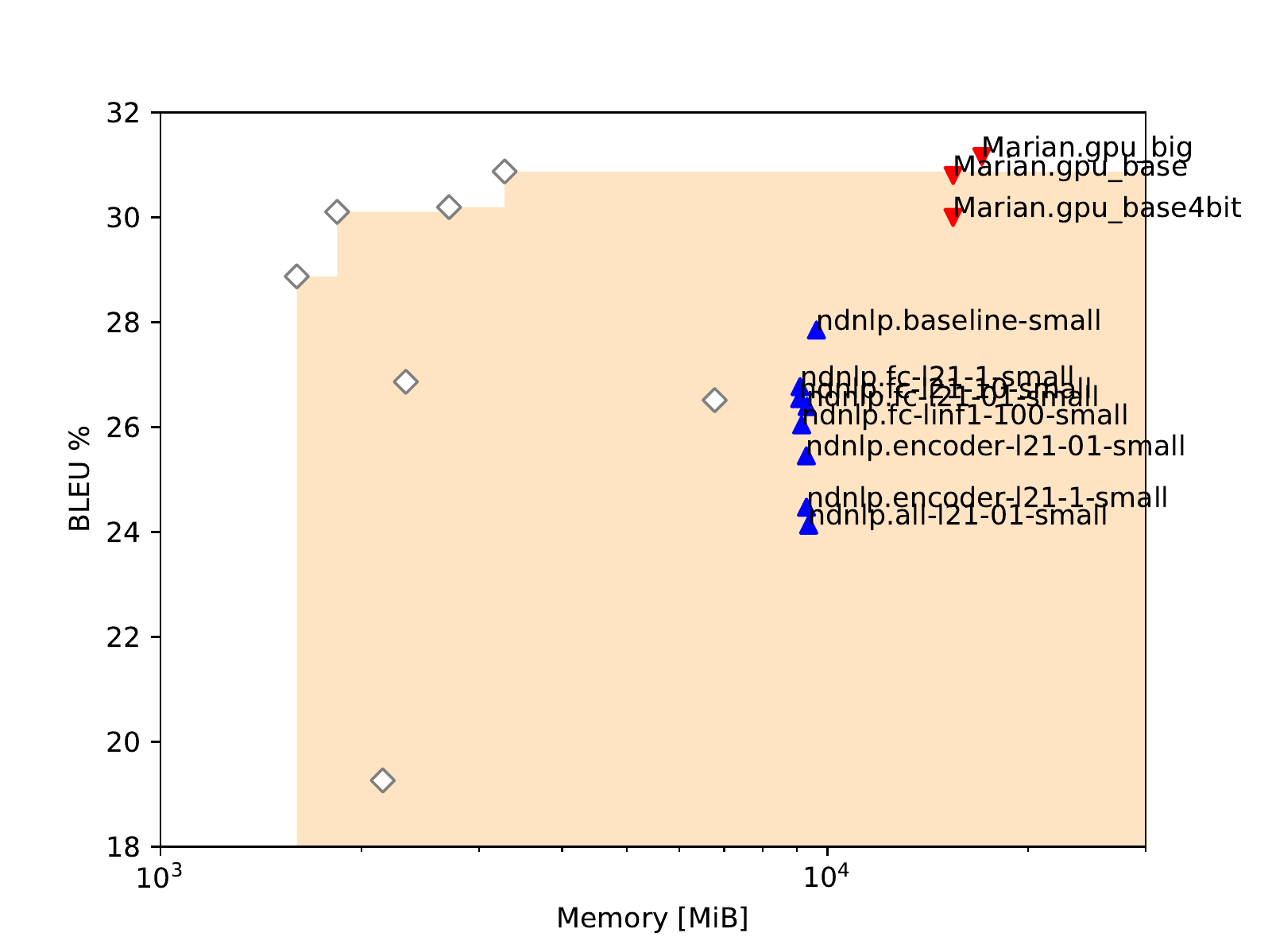} }
  \caption{Time and memory vs. accuracy measured by BLEU on the newstest2015 set, calculated on both CPU and GPU. White diamonds ($\diamond$) represent the results in the previous campaign. Orange areas show regions dominated by some Pareto frontier systems.}
  \label{fig:results}
\end{figure*}

\subsection{Baseline Systems}

Two baseline systems were prepared:
\begin{description}
\item[Echo:] Just send the input back to the output.
\item[Base:] A baseline system using attentional LSTM-based encoder-decoders with attention \cite{bahdanau15alignandtranslate}.
\end{description}

\subsection{Submitted Systems}

Two teams, Team Marian and Team Notre Dame submitted to the shared task, and we will summarize each below.

\subsubsection{Team Marian}

Team Marian's submission \cite{kim19marian} was based on their submission to the shared task the previous year, consisting of Transformer models optimized in a number of ways \cite{junczys-dowmunt-etal-2018-marian}.
This year, they made a number of improvements.
Improvements were made to teacher-student training by (1) creating more data for teacher-student training using backward, then forward translation, (2) using multiple teachers to generate better distilled data for training student models.
In addition, there were modeling improvements made by (1) replacing simple averaging in the attention layer with an efficiently calculable ``simple recurrent unit,'' (2) parameter tying between decoder layers, which reduces memory usage and improves cache locality on the CPU.
Finally, a number of CPU-specific optimizations were performed, most notably including 8-bit matrix multiplication along with a flexible quantization scheme.

\subsubsection{Team Notre Dame}

Team Notre Dame's submission \cite{murray19notredame} focused mainly on memory efficiency.
They did so by performing ``Auto-sizing'' of the transformer network, applying block-sparse regularization to remove columns and rows from the parameter matrices.

\subsection{Results}
\label{sec:sharedtaskresults}

\begin{figure}[t]
  \centering
  \includegraphics[width=0.99\linewidth]{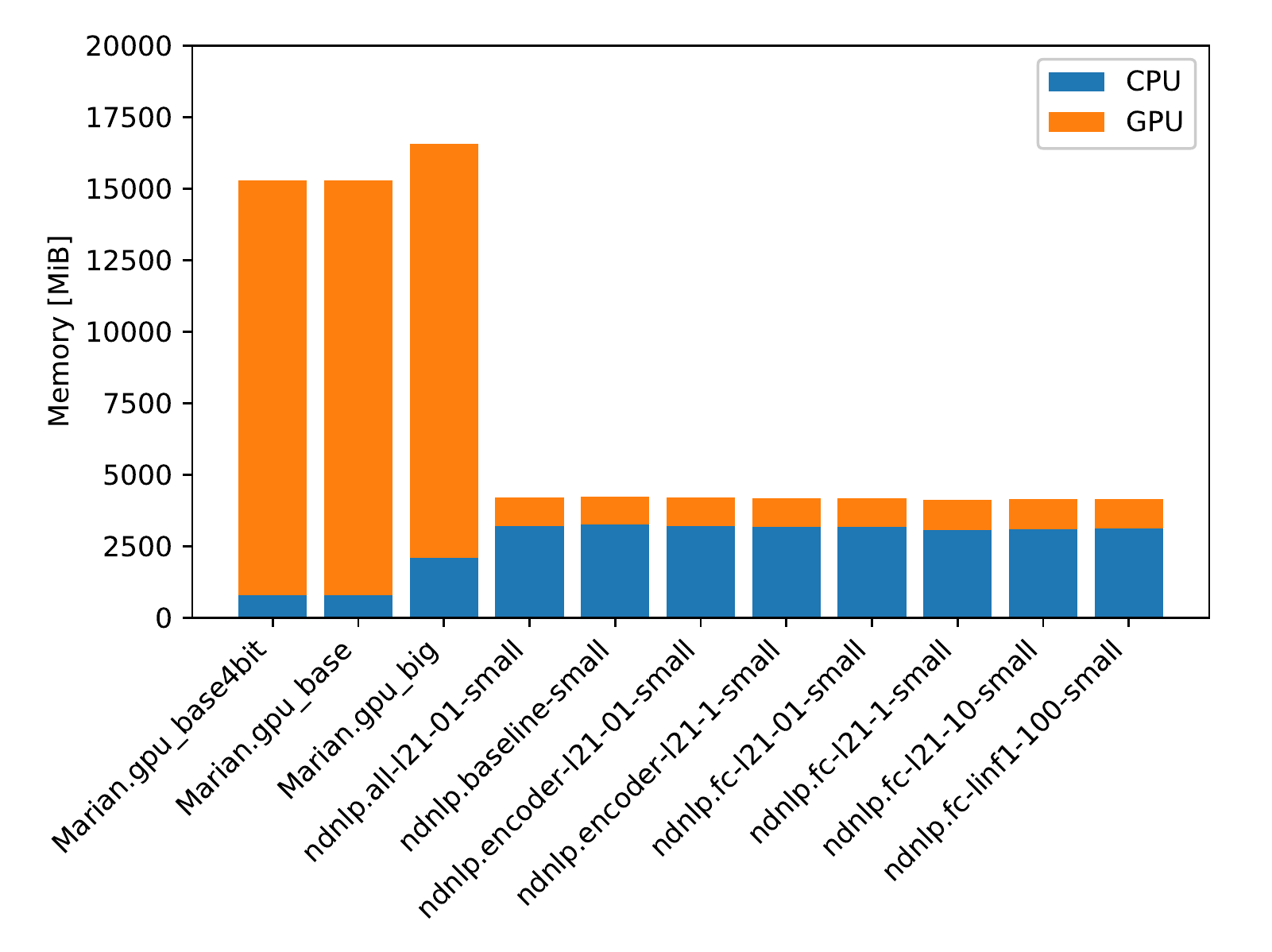}
  (a) empty
  \includegraphics[width=0.99\linewidth]{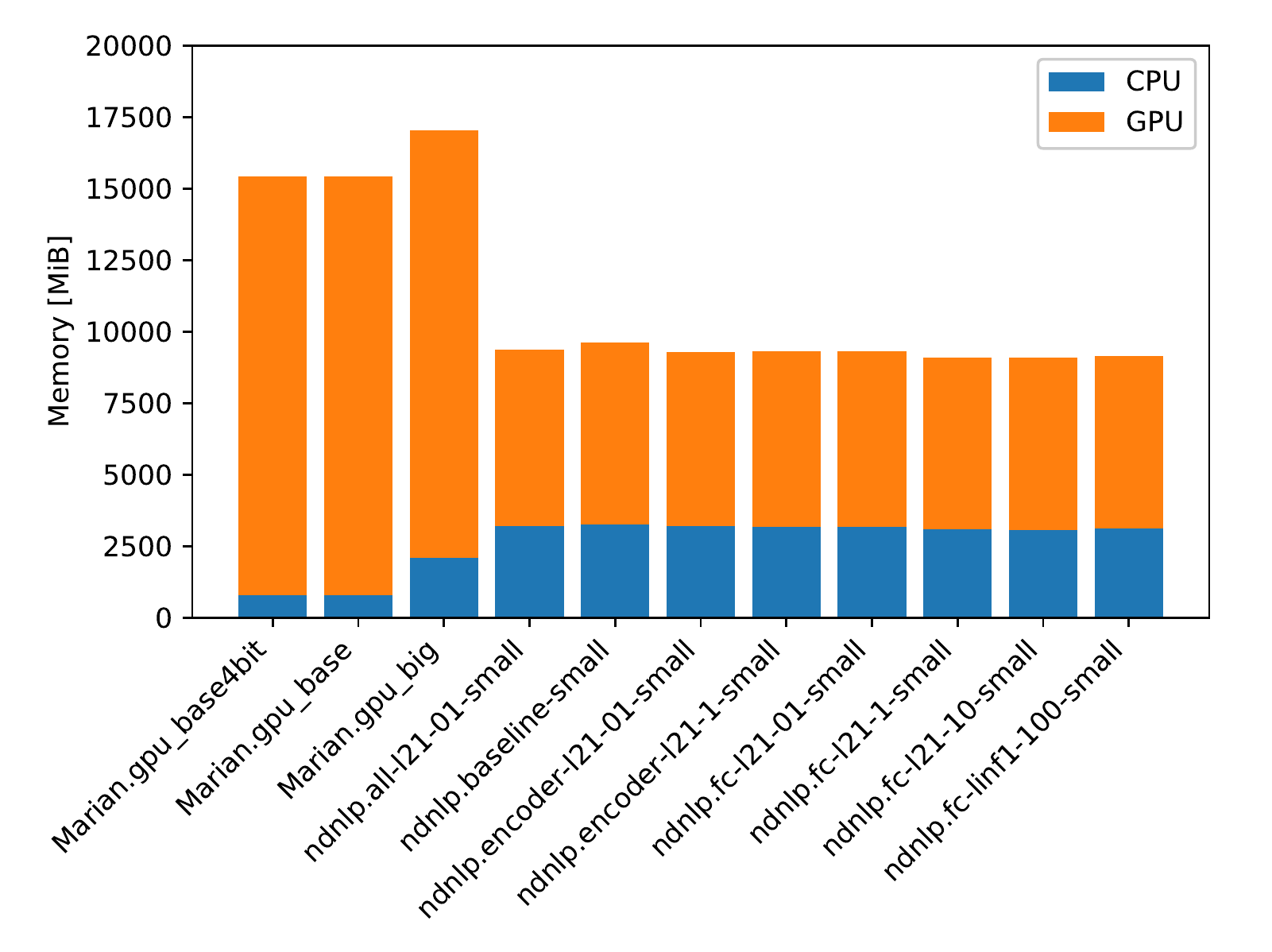}
  (b) newstest2015
  \caption{Memory consumption for each GPU system.}
  \label{fig:gpumem}
\end{figure}

A brief summary of the results of the shared task (for newstest2015) can be found in Figure \ref{fig:results}, while full results tables for all of the systems can be found in Appendix \ref{sec:fullresults}.
From this figure we can glean a number of observations.

For the CPU systems, all submissions from the Marian team clearly push the Pareto frontier in terms of both time and memory.
In addition, the Marian systems also demonstrated a good trade-off between time/memory and accuracy.

For the GPU systems, all systems from the Marian team also outperformed other systems in terms of the speed-accuracy trade-off.
However, the Marian systems had larger memory consumption than both Notre Dame systems, which specifically optimized for memory efficiency, and all previous systems.
Interestingly, each GPU system by the Marian team shares almost the same amount of GPU memory as shown in Table \ref{tab:memory_gpu} and Figure \ref{fig:gpumem}(b).
This may indicate that the internal framework of the Marian system tries to reserve enough amount of the GPU memory first, then use the acquired memory as needed by the translation processes.

On the other hand, we can see that the Notre Dame systems occupy only a minimal amount of GPU memory, as the systems use much smaller amounts on the empty set (Figure \ref{fig:gpumem}(a)).
These different approaches to constant or variable size memory consumption may be based on different underlying perspectives of ``memory efficiency,'' and it may be difficult to determine which policy is better without knowing the actual environment in which a system will be used.

\section{Conclusion}

This paper summarized the results of the Third Workshop on Neural Generation and Translation, where we saw a number of research advances.
Particularly, this year introduced a new document generation and translation task, that tested the efficacy of systems for both the purposes of translation and generation in a single testbed.

\section*{Acknowledgments}

We thank Apple and Google for their monetary support of student travel awards for the workshop, and AWS for its gift of AWS credits (to Graham Neubig) that helped support the evaluation.

\bibliography{myabbrv,emnlp-ijcnlp-2019,wngt19}
\bibliographystyle{acl_natbib}

\appendix
\section{Full Shared Task Results}
\label{sec:fullresults}

For completeness, in this section we add tables of the full shared task results. These include the full size of the image file for the translation system (Table \ref{tab:image_file_sizes}), the comparison between compute time and evaluation scores on CPU (Table \ref{tab:time_and_eval_cpu}) and GPU (Table \ref{tab:time_and_eval_gpu}), and the comparison between memory and evaluation scores on CPU (Table \ref{tab:memory_cpu}) and GPU (Table \ref{tab:memory_gpu}).

\begin{table*}[t]
\begin{center}
\caption{Image file sizes of submitted systems.}
\label{tab:image_file_sizes}
\begin{tabular}{|ll|r|}
\hline
Team & System & Size [MiB] \\
\hline
Marian     & cpu\_base2            &  135.10 \\
           & cpu\_medium1          &  230.22 \\
           & cpu\_small2           &   93.00 \\
           & cpu\_tiny11           &   91.61 \\
           & gpu\_base4bit         &  322.68 \\
           & gpu\_base             &  452.59 \\
           & gpu\_big              &  773.27 \\
Notre Dame & all-l21-01-small     & 2816.63 \\
           & baseline-small       & 2845.04 \\
           & encoder-l21-01-small & 2813.67 \\
           & encoder-l21-1-small  & 2779.60 \\
           & fc-l21-01-small      & 2798.94 \\
           & fc-l21-1-small       & 2755.80 \\
           & fc-l21-10-small      & 2755.76 \\
           & fc-linf1-100-small   & 2759.10 \\
\hline
\end{tabular}
\end{center}
\end{table*}
\begin{table*}[t]
\begin{center}
\caption{Time consumption and MT evaluation metrics (CPU systems).}
\label{tab:time_and_eval_cpu}
\resizebox{\textwidth}{!}{%
\begin{tabular}{|l|ll|rr|rr|}
\hline
\multirow{2}{*}{Dataset} &
\multirow{2}{*}{Team} &
\multirow{2}{*}{System} &
\multicolumn{2}{|c|}{Time Consumption [s]} &
\multirow{2}{*}{BLEU \%} &
\multirow{2}{*}{NIST} \\
& & & Total & Diff & & \\
\hline
Empty
  & Marian
    & cpu\_base2   &   4.97 &    --- &   --- &   --- \\
  & & cpu\_medium1 &   6.03 &    --- &   --- &   --- \\
  & & cpu\_small2  &   4.67 &    --- &   --- &   --- \\
  & & cpu\_tiny11  &   4.71 &    --- &   --- &   --- \\
\hline
newstest2014
  & Marian
    & cpu\_base2   &  57.52 &  52.55 & 28.04 & 7.458 \\
  & & cpu\_medium1 & 181.81 & 175.78 & 28.58 & 7.539 \\
  & & cpu\_small2  &  28.32 &  23.64 & 26.97 & 7.288 \\
  & & cpu\_tiny11  &  22.26 &  17.56 & 26.38 & 7.178 \\
\hline
newstest2015
  & Marian
    & cpu\_base2   &  45.01 &  40.04 & 30.74 & 7.607 \\
  & & cpu\_medium1 & 139.06 & 133.03 & 31.32 & 7.678 \\
  & & cpu\_small2  &  22.64 &  17.97 & 29.57 & 7.437 \\
  & & cpu\_tiny11  &  18.08 &  13.37 & 28.92 & 7.320 \\
\hline
\end{tabular}
}
\end{center}
\end{table*}

\begin{table*}[t]
\begin{center}
\caption{Time consumption and MT evaluation metrics (GPU systems).}
\label{tab:time_and_eval_gpu}
\resizebox{\textwidth}{!}{%
\begin{tabular}{|l|ll|rr|rr|}
\hline
\multirow{2}{*}{Dataset} &
\multirow{2}{*}{Team} &
\multirow{2}{*}{System} &
\multicolumn{2}{|c|}{Time Consumption [s]} &
\multirow{2}{*}{BLEU \%} &
\multirow{2}{*}{NIST} \\
& & & Total & Diff & & \\
\hline
Empty
  & Marian
    & gpu\_base4bit        &  9.77 &   --- &   --- &   --- \\
  & & gpu\_base            &  2.69 &   --- &   --- &   --- \\
  & & gpu\_big             &  5.22 &   --- &   --- &   --- \\
  & Notre Dame
    & all-l21-01-small     &  7.92 &   --- &   --- &   --- \\
  & & baseline-small       &  8.13 &   --- &   --- &   --- \\
  & & encoder-l21-01-small &  7.96 &   --- &   --- &   --- \\
  & & encoder-l21-1-small  &  7.89 &   --- &   --- &   --- \\
  & & fc-l21-01-small      &  7.85 &   --- &   --- &   --- \\
  & & fc-l21-1-small       &  7.56 &   --- &   --- &   --- \\
  & & fc-l21-10-small      &  7.57 &   --- &   --- &   --- \\
  & & fc-linf1-100-small   &  7.61 &   --- &   --- &   --- \\
\hline
newstest2014
  & Marian
    & gpu\_base4bit        & 12.42 &  2.66 & 27.50 & 7.347 \\
  & & gpu\_base            &  4.22 &  1.53 & 28.00 & 7.449 \\
  & & gpu\_big             &  7.09 &  1.88 & 28.61 & 7.534 \\
  & Notre Dame
    & all-l21-01-small     & 36.80 & 28.89 & 21.60 & 6.482 \\
  & & baseline-small       & 36.07 & 27.95 & 25.28 & 7.015 \\
  & & encoder-l21-01-small & 35.82 & 27.86 & 23.20 & 6.725 \\
  & & encoder-l21-1-small  & 35.41 & 27.52 & 22.06 & 6.548 \\
  & & fc-l21-01-small      & 35.76 & 27.91 & 24.07 & 6.869 \\
  & & fc-l21-1-small       & 34.46 & 26.90 & 23.97 & 6.859 \\
  & & fc-l21-10-small      & 34.00 & 26.43 & 23.87 & 6.852 \\
  & & fc-linf1-100-small   & 34.46 & 26.84 & 23.80 & 6.791 \\
\hline
newstest2015
  & Marian
    & gpu\_base4bit        & 11.97 &  2.20 & 29.99 & 7.504 \\
  & & gpu\_base            &  3.98 &  1.29 & 30.79 & 7.595 \\
  & & gpu\_big             &  6.80 &  1.58 & 31.15 & 7.664 \\
  & Notre Dame
    & all-l21-01-small     & 30.21 & 22.29 & 24.13 & 6.670 \\
  & & baseline-small       & 30.47 & 22.35 & 27.85 & 7.224 \\
  & & encoder-l21-01-small & 29.67 & 21.71 & 25.45 & 6.869 \\
  & & encoder-l21-1-small  & 29.93 & 22.04 & 24.47 & 6.734 \\
  & & fc-l21-01-small      & 29.80 & 21.95 & 26.39 & 7.053 \\
  & & fc-l21-1-small       & 28.42 & 20.87 & 26.77 & 7.093 \\
  & & fc-l21-10-small      & 28.63 & 21.06 & 26.54 & 7.051 \\
  & & fc-linf1-100-small   & 28.91 & 21.29 & 26.04 & 6.971 \\
\hline
\end{tabular}
}
\end{center}
\end{table*}

\begin{table*}[t]
\begin{center}
\caption{Peak memory consumption (CPU systems).}
\label{tab:memory_cpu}
\begin{tabular}{|l|ll|rrr|}
\hline
\multirow{2}{*}{Dataset} &
\multirow{2}{*}{Team} &
\multirow{2}{*}{System} &
\multicolumn{3}{|c|}{Memory [MiB]} \\
& & & Host & GPU & Both \\
\hline
Empty
  & Marian
    & cpu\_base2   & 336.30 & --- & 336.30 \\
  & & cpu\_medium1 & 850.76 & --- & 850.76 \\
  & & cpu\_small2  & 229.53 & --- & 229.53 \\
  & & cpu\_tiny11  & 227.73 & --- & 227.73 \\
\hline
newstest2014
  & Marian
    & cpu\_base2   & 523.93 & --- & 523.93 \\
  & & cpu\_medium1 & 850.98 & --- & 850.98 \\
  & & cpu\_small2  & 276.30 & --- & 276.30 \\
  & & cpu\_tiny11  & 260.86 & --- & 260.86 \\
\hline
newstest2015
  & Marian
    & cpu\_base2   & 523.12 & --- & 523.12 \\
  & & cpu\_medium1 & 850.95 & --- & 850.95 \\
  & & cpu\_small2  & 275.76 & --- & 275.76 \\
  & & cpu\_tiny11  & 260.09 & --- & 260.09 \\
\hline
\end{tabular}
\end{center}
\end{table*}

\begin{table*}[t]
\begin{center}
\caption{Peak memory consumption (GPU systems).}
\label{tab:memory_gpu}
\begin{tabular}{|l|ll|rrr|}
\hline
\multirow{2}{*}{Dataset} &
\multirow{2}{*}{Team} &
\multirow{2}{*}{System} &
\multicolumn{3}{|c|}{Memory [MiB]} \\
& & & Host & GPU & Both \\
\hline
Empty
  & Marian
    & gpu\_base4bit        &  788.67 & 14489 & 15277.67 \\
  & & gpu\_base            &  791.82 & 14489 & 15280.82 \\
  & & gpu\_big             & 2077.75 & 14489 & 16566.75 \\
  & Notre Dame
    & all-l21-01-small     & 3198.89 &   991 &  4189.89 \\
  & & baseline-small       & 3261.73 &   973 &  4234.73 \\
  & & encoder-l21-01-small & 3192.87 &  1003 &  4195.87 \\
  & & encoder-l21-1-small  & 3164.45 &  1003 &  4167.45 \\
  & & fc-l21-01-small      & 3160.32 &   999 &  4159.32 \\
  & & fc-l21-1-small       & 3069.05 &  1049 &  4118.05 \\
  & & fc-l21-10-small      & 3092.01 &  1057 &  4149.01 \\
  & & fc-linf1-100-small   & 3116.35 &  1037 &  4153.35 \\
\hline
newstest2014
  & Marian
    & gpu\_base4bit        &  781.83 & 14641 & 15422.83 \\
  & & gpu\_base            &  793.07 & 14641 & 15434.07 \\
  & & gpu\_big             & 2078.78 & 14961 & 17039.78 \\
  & Notre Dame
    & all-l21-01-small     & 3199.95 &  9181 & 12380.95 \\
  & & baseline-small       & 3285.20 &  9239 & 12524.20 \\
  & & encoder-l21-01-small & 3194.96 &  9169 & 12363.96 \\
  & & encoder-l21-1-small  & 3164.86 &  9119 & 12283.86 \\
  & & fc-l21-01-small      & 3160.80 &  9155 & 12315.80 \\
  & & fc-l21-1-small       & 3070.67 &  9087 & 12157.67 \\
  & & fc-l21-10-small      & 3068.12 &  9087 & 12155.12 \\
  & & fc-linf1-100-small   & 3119.96 &  9087 & 12206.96 \\
\hline
newstest2015
  & Marian
    & gpu\_base4bit        &  783.05 & 14641 & 15424.05 \\
  & & gpu\_base            &  789.34 & 14641 & 15430.34 \\
  & & gpu\_big             & 2077.42 & 14961 & 17038.42 \\
  & Notre Dame
    & all-l21-01-small     & 3198.86 &  6171 &  9369.86 \\
  & & baseline-small       & 3266.11 &  6359 &  9625.11 \\
  & & encoder-l21-01-small & 3193.13 &  6103 &  9296.13 \\
  & & encoder-l21-1-small  & 3166.56 &  6135 &  9301.56 \\
  & & fc-l21-01-small      & 3160.88 &  6159 &  9319.88 \\
  & & fc-l21-1-small       & 3079.92 &  6017 &  9096.92 \\
  & & fc-l21-10-small      & 3069.32 &  6015 &  9084.32 \\
  & & fc-linf1-100-small   & 3130.95 &  6015 &  9145.95 \\
\hline
\end{tabular}
\end{center}
\end{table*}

\end{document}